\documentclass{article}

 \usepackage[preprint]{nips_2018}

\usepackage[utf8]{inputenc} 
\usepackage[T1]{fontenc}    
\usepackage{hyperref}       
\usepackage{url}            
\usepackage{booktabs}       
\usepackage{amsfonts}       
\usepackage{nicefrac}       
\usepackage{microtype}      
\usepackage[utf8]{inputenc}
\usepackage{comment}
\usepackage{color}
\usepackage{multirow}
\usepackage{multicol}
\usepackage{amsmath}
\usepackage{graphicx}
\usepackage{caption}
\usepackage{tikz}
\usepackage{enumitem}

\newcommand{\instr}[1]{{\fontfamily{cmtt}\selectfont#1}}

\newcommand{\ignore}[1]{}

\title{Human Instruction-Following with Deep Reinforcement Learning\\ via Transfer-Learning from Text}

\author{
  Felix Hill\thanks{\texttt{felixhill@google.com}}, So\v na Mokr\'a,    Nathaniel Wong, Tim Harley   \\
  DeepMind, London\\
}

\begin{document}

\maketitle

\begin{abstract}
Recent work has described neural-network-based agents that are trained with reinforcement learning (RL) to execute language-like commands in simulated worlds, as a step towards an intelligent agent or robot that can be instructed by human users. However, the optimisation of multi-goal motor policies via deep RL from scratch requires many episodes of experience. Consequently, instruction-following with deep RL typically involves language generated from templates (by an environment simulator), which does not reflect the varied or ambiguous expressions of real users. Here, we propose a conceptually simple method for training instruction-following agents with deep RL that are robust to natural human instructions. By applying our method with a state-of-the-art pre-trained text-based language model (BERT), on tasks requiring agents to identify and position everyday objects relative to other objects in a naturalistic 3D simulated room, we demonstrate substantially-above-chance zero-shot transfer from synthetic template commands to natural instructions given by humans. Our approach is a general recipe for training any deep RL-based system to interface with human users, and bridges the gap between two research directions of notable recent success: agent-centric motor behavior and text-based representation learning. 
\end{abstract}

\section{Introduction}

Developing machines that can follow natural human commands, particularly those pertaining to an environment shared by both machine and human, is a long-standing and elusive goal of AI~\citep{winograd1972understanding}. Recent work has applied deep reinforcement learning (RL) methods to this challenge, where a neural-network-based agent is optimized to process language input, perceive its surroundings and execute appropriate movements jointly \citep{oh2017zero,hermann2017grounded,chaplot2018gated,zhong2019rtfm}. Deep RL promises a way to deal flexibly with the complexity of the physical, visual and linguistic world without relying on (potentially brittle) hand-crafted features, rules or policies. Nevertheless, the cost of this flexibility is the large number of environment interactions (samples) required for a network to learn behaviour policies from raw experience. To make the approach feasible, many studies thus employ a synthetic language that is generated on demand from templates by the environment simulator~\citep{chevalier2018babyai,jiang2019language,yu2018interactive,yu2018guided,zhong2019rtfm}. Attempts to integrate deep RL with natural language typically employ less realistic grid-like environments~\citep{misra2017mapping} or grant agents privileged global observations and gold-standard action sequences to make learning more tractable~\citep{misra2018mapping}. 

\begin{figure}
    \centering
    \includegraphics[width=8.5cm,keepaspectratio]{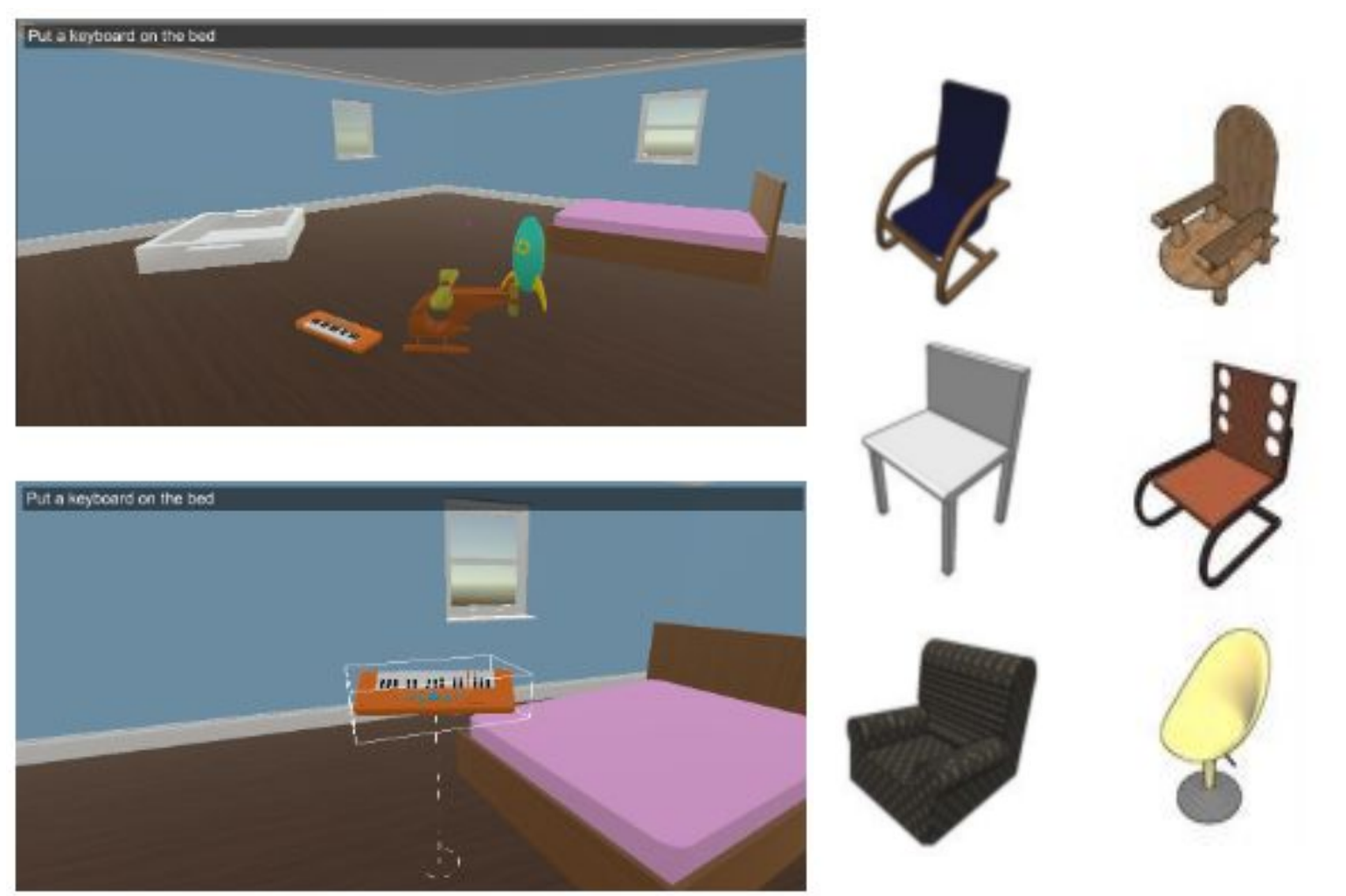}
    \caption{\label{fig:room} \textbf{Left:} stills of the \textit{putting task} from the agent's perspective, involving three randomly coloured and positioned moveable objects, plus a pink bed and a white tray. During evaluation, the agent receives an instruction given by a human (e.g. \emph{``Drop A Casio onto the bed"}), identify the intended object of reference, move towards it and lift it, rotating if necessary, and then lower it into the specified location. \textbf{Right:} example ShapeNet models from the class `chair` used in the reference task, demonstrating the perceptual variation which in turn drives substantial linguistic variation across the human testing instructions. }
\end{figure}

Here, we propose an alternative learning-based recipe for training deep RL agents that are robust to a natural human commands, which we call SHIFTT (Simulation-to-Human Instruction Following via Transfer from Text). In this approach, agents are first endowed with powerful pretrained language encoders and then their visual processing and behaviour policies are optimised using conventional deep RL methods to respond to synthetic, template-based language. Finally, we evaluate the agents on their ability to execute instructions from human testers that are (in theory) semantically-equivalent, but can be superficially quite distinct from those experienced by the agent during training. 

We demonstrate the effectiveness of this approach in the context of a 3D room containing object models from the ShapeNet dataset~\citep{chang2015shapenet}. Using 3D models based on everyday objects allows us to solicit from human testers diverse ways of instructing and/or referring to things, and hence to probe the agents' robustness to this diversity. Unsurprisingly, we find that agents trained with conventional template-based commands do not adapt well to the natural (keyboard-typed) commands of the testers. With SHIFTT, however, agents can satisfy human instructions with substantially above-chance accuracy, which shows that powerful language models can support a notable degree of zero-shot policy generalization. Indeed, on the two tasks we consider in this work, our agent, trained only in simulation, performs as well as naive human game operators at the task of executing the noisy instructions provided by other humans. 

To better analyze the generalization that supports this robustness, we consider different ways of incorporating pre-training language models into a deep RL agent, and probe each condition with specific synthetic deviations from their template-based training instructions. We find that pre-trained language encoders based on non-BERT (context-free) word embeddings can support a degree of generalization driven by lexical similarity (executing \instr{Find a vehicle} when trained to \instr{Find a car}), but not phrasal equivalence (failing at \instr{Put a plate on the container} when trained to \instr{Put the dish on the tray}). In contrast, methods that integrate powerful contextual word representations (i.e. BERT) support both types of generalization. We note further boosts to test robustness when these pretrained language encoders are complemented with additional learned self-attention layers (tuned to the agent's environmental objectives). Ablation experiments isolate the role of \emph{WordPiece tokenization}~\citep{schuster2012japanese} in robustness to typed human instructions. This motivates the addition of \emph{typo noise} as a critical component of the training pipeline for the highest-performing agents. 


\paragraph{The SHIFTT method}

\begin{figure*}
    \centering
    \makebox[\textwidth][c]{\includegraphics[width=1.2\textwidth]{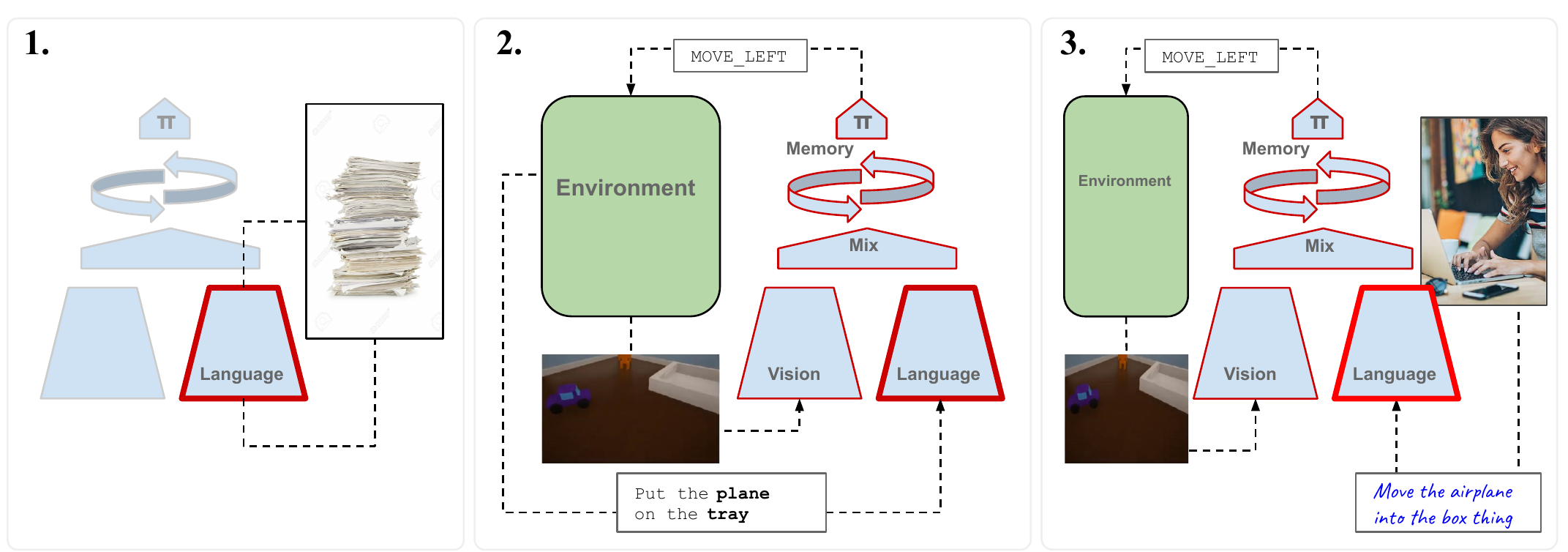}}
    \caption{\label{fig:shiftt} The three stages of the SHIFTT method. During RL training (2.) the influence of semantic knowledge induced via text-based pre-training (1.) is gradually distributed across the agent's policy network, yielding an agent robust to human instructions (3.).}
\end{figure*}

In the general case, our proposed recipe (Fig~\ref{fig:shiftt}) involves the following steps:

\begin{enumerate}
\item
Train or acquire a pretrained language model $L$.
\item
Construct a language-dependent object manipulation task. Define global set of human-nameable objects $G$ and a class of binary spatial relations $S$. For each $s \in S$, write a reward function $f_s$, such that $f_s(o_1, o_2) > 0 $ $ \forall o_1, o_2 \in G$ iff $o_1, o_2$ are in spatial relation $s$. 

For each episode:
\begin{enumerate} 
\item Sample a subset $G' \subset G, \|g\| > 2$ of unique objects, and individual objects $o_1, o_2 \in G'$.
\item Sample a spatial relation $s \in S$. 
\item Construct an instruction $i$ according to the template \instr{Put the } $w_{o_1} w_s$ \instr{the}  $w_{o_2}$, where $w_x$ is the everyday name for object or relation $x$. 
\item Spawn all $g \in G'$ and an agent at random positions and orientations in the environment. Append $i$ to agent environment observations.
\end{enumerate}
\item Train the agent using RL to maximize the expected cumulative rewards from $f$ based on per-timestep visual observations and instructions $i$ encoded by $L$.

\item 
Have human testers interact with episodes in the environment and pose instructions to the agent (with the macro-objective of having relations $s \in S$ realized for some $o_1, o_2 \in G'$ each episode). Evaluate the performance of the agent in response to these instructions. 
\end{enumerate}

In this work, we apply SHIFTT in a simulated room in the Unity\footnote{Unity. http://unity3d.com.} game engine. Note that there is no reason, in theory at least, why the recipe could not apply to robot-learning for object manipulation (e.g. ~\citet{johannink2019residual}).

\section{Architectures for instruction-following agents}
Since the object of our study is instruction-following, we consider an agent architecture that is as conventional as possible in aspects not related to language processing. It consists of standard components for computing visual representations of pixels, embedding text strings and predicting actions contingent on these and memory of past inputs.\footnote{See Supplement for details not included here.}

\paragraph{Visual processing} The visual observations received by the agent at each timestep are $96 \times 72 \times 3$ real-valued tensors, which are processed by a 3-layer residual convnet. 

\paragraph{Memory core} The output of the visual processing is combined with language information according to a particular encoding strategy, as described below. In all conditions, some combination of vision and language input at each timestep passes into a LSTM memory with hidden dim. $128$. 

\paragraph{Action and value prediction} The state of the memory core at each timestep is passed through a linear layer and softmax to compute a distribution over 26 actions. Independently, the memory state is passed to a linear layer to yield a scalar value prediction. 

\paragraph{Training algorithm} The agent is trained using an importance-weighted actor-critic algorithm with a central learner and distributed actors~\citep{espeholt2018impala}. 

\subsection{Language encoding} 

The agent must process both string observations in the simulated environment (during training) and human instructions, also encoded as strings (during evaluation). Its representations of language must be combined with visual information to make decisions about how to act. We compare various different ways of achieving this encoding. For methods that transfer knowledge from unsupervised text-based learning, we take weights from the well-known BERT model ~\citep{devlin2018bert}, specifically the  uncased BERT\textsubscript{BASE} model made available by the authors.\footnote{Available at \url{https://tfhub.dev/google/bert_uncased_L-12_H-768_A-12/1}}

\textbf{BERT + mean pool}~~For a given input of $w$ words, BERT\textsubscript{BASE} returns $w$ context-dependent (sub)-word representations of $768$ units. In this condition, a mean pooling operation is applied over the $w$ dimension to yield a single representation of dimension $768$, which is concatenated with the flattened output of the visual processing module. This multi-modal representation is passed through a single layer MLP with $tanh$ activation and output dimension $128$ before entering the memory core of the agent. We apply the standard BERT\textsubscript{BASE} WordPiece vocabulary (of size 30,000). Note that WordPiece encodes language in terms of \textit{subwords} (a mix of characters, common word chunks, morphemes and words) rather than the word-level vocabulary applied in more traditional neural language models.  

\textbf{BERT + self-attention layer}~~When applying BERT to text classification, performance can sometimes be improved by \emph{fine-tuning} the weights in the BERT encoder to suit the task. Because of the large number of gradient updates required to learn a complex behaviour policy, fine-tuning the BERT weights in this way would cause substantial overfitting to the synthetic environment language. We therefore keep the BERT weights frozen, but experiment with an additional learned self-attention layer~\citep{vaswani2017attention} to create a language encoder whose bottom layers are pretrained and whose final layer optimizes these representations to the present environment or tasks. This additional layer has $4$ attention heads, and uses $64$ dimensional key and value embeddings.

\textbf{BERT+ cross-modal self-attention}~~We also consider a cross-modal self-attention layer of the type suggested by e.g.~\cite{tsai2019multimodal,lu2019vilbert}, which in our application provides an explicit pathway for the agent to bind visual experience to specific (contextual) word representations. In this case, we treat each of the output channels of the visual processing network as word-like entities, by passing them through a linear layer of output size $768$ to match the BERT output. The embeddings for all words and the visual channels are then processed with a single self-attention layer, whose weights are again learned with other agent parameters. Note that, when applied across modalities, a Transformer self-attention layer is more expressive than prior attention-based operations for fusing vision and language in agents ~\cite{chaplot2018gated}, since it permits interactions at the (contextualized) word-level rather than the entire instruction, because it includes multiple heads, and because the key-query-value mechanism allows the model to learn both content-based and key-based lookup processes. 

\textbf{Pretrained (sub)-word embeddings}~~Language classification experiments with BERT show the value of highly context-dependent (sub)-word representations, but transfer in that context is also possible with more conventional (context-independent) word embeddings \citep{collobert2008unified}. To measure the effect of this distinction, we consider a simpler encoder based on the (context independent) input (sub-)word embeddings from BERT, which are also of dimension $768$.\footnote{These embeddings capture lexical similarity much like conventional word embeddings; a cosine metric produces a Spearman correlation of $0.49$ with human ratings from the Simlex-999 dataset~\citep{hill2015simlex}.} Taking a mean of these context-independent vectors would yield a word-order-invariant representation of language. We therefore process them with single-layer transformer with $4$ attention heads. As in \emph{BERT + mean pool}, this output is averaged, reduced by a single-layer MLP (from $768$ to $128$ units) and passed to the agent's core memory.

\textbf{Typo noise}~~Finally, in one condition we introduce typo noise when training the agent (SHIFTT stage 2). The noise function that we apply simply replaces each key character of the template language strings produced by the environment with that of an adjacent character on a standard keyboard with a probability of $0.01$. This encourages the agent to learn to compensate, during the training process, to potential typographical errors of human operators. No modifications were made to the evaluation stimuli. Much previous work applies typing noise for language classifier robustness; see e.g. \citep{pruthi2019combating} for a recent survey.

We compare to various baselines designed to isolate specific components of these encoders. 

\textbf{Random mean pool}~~As a direct baseline for~\emph{BERT mean pool}, we consider an identical architecture but in which the contextual BERT embeddings are replaced by fixed random (subword-specific) vectors (also of dimension $768$). The weights in the rest of the agent network are trained as in other conditions. We note that random sentence vectors are a competitive baseline on many language classification tasks~\citep{wieting2019no}.

\textbf{Word-level and WordPiece transformers}~~In addition to pre-trained weights, a potentially important aspect of encoders based on BERT is WordPiece tokenization, which can afford greater ability to make sense of typos or rare words with familiar morphemes than in a more conventional word-level encoder. To isolate the effect of tokenization from that of pretrained weights, we compared two further encoders. In one, we split all input strings by whitespace and hash each word string to a unique index representing an input to a single-layer transformer with $4$ attention heads and embedding size $768$ (chosen to match BERT\textsubscript{BASE}), the output of which is processed identically to the Random mean pool condition. We contrast this with an otherwise identical condition in which the WordPiece tokenization from BERT is applied rather than splitting on white space.

\textbf{Human performance estimate}~~As a point of comparison, we recruited a set of five human testers, separate from those that produced the original instructions, and subjected them to a similar evaluation procedure as the agent for twenty episodes of the reference and putting task. Observing these episodes, we found the reasons for human failure on a given episode to include unclear or ambiguous instructions (such as \emph{Put a toy on the bed} when any object could conceivably be a toy), objects with ambiguous appearance (e.g. the car and bus look similar), and (rarely) a failure of control or object manipulation.

\section{Experiments}

We experiment with a \textbf{reference task}, which focuses on object identification, and a \textbf{putting task} which focuses on object relations and discrete object control. In both tasks the locations of all objects and the initial position and orientation of the agent are chosen at random on floating point scales, so it is highly unlikely that any two episodes are spatially identical (whether during training or evaluation). The action set consists of movements ({\small {\tt move-}$\{${\tt forward,back,left,right}$\}$}),
turns ({\small {\tt turn-}$\{${\tt up,down,left,right}$\}$}), and object pick-up and hand control with $4$ DoF. The placement of objects is assisted by a visible vertical line. See Supplement for a full description.

\subsection{Determining reference of ShapeNet objects}

In an episode of the reference task, the environment selects two movable objects at random from a global object set, and generates a language instruction of the form \instr{Find a X}, where \instr{X} is the correct name for (exactly) one of those objects. To achieve reward, the agent must locate the \instr{X}, lift it more than 1m above the ground and keep it there for 5 time steps. If the reward function in the environment determines that these conditions are met, a positive reward of $+1$ is emitted and the episode ends. If a lifting takes place with the incorrect object, the episode also ends with reward $0$. For a global set of recognisable objects, we use the \emph{ShapeNet} dataset \citep{chang2015shapenet}, which contains over 12,000 3D rendered models. For performance reasons we discard models with more than 8,000 vertices or an OBJ file size greater than 1 MB and consider only those tagged into a synset in the WordNet taxonomy. We take models from the ShapeNetSem slice of the data, and use the name of the first lemma of that synset as a proxy for the name of the object. From these names, we selected 80 for the environment, each referring to a set of object models with a minimum of 12 exemplars. See Supplement for details.

\begin{table}
\centering
\small
\begin{tabular}{l|l|l}
\textbf{Template lang.} & \textbf{}                & \textbf{Natural} \\
\textbf{(training)} & \textbf{Synonym}                & \textbf{reference} \\ \hline
\multirow{3}{*}{Find a flag}              & \multirow{3}{*}{Find a}  & Find \textit{the indian flag}          \\
                                          &                               & Find \textit{a flag}                   \\
                                          &          \textbf{banner}                         & Find \textit{the flag.}                \\
                                         &                      & Find \textit{the cardboard.}                \\\hline
\multirow{3}{*}{Find a pillow}            & \multirow{3}{*}{Find a} & Find \textit{a pillows}                \\
                                          &                             & Find \textit{a cushion}                \\
                                          &     \textbf{cushion}                              & Find \textit{a paper} 
 \\
                                          &                            & Find \textit{a set of cushions} 
\end{tabular}

\caption{\label{tab:liftexp} Example training/test instructions in the reference task.}  

\end{table}

For all language encoding strategies described above, the agent was able to learn the task, and a well-trained policy completed episodes in an average of $\approx 20$ timesteps with accuracy around 90\%. The failure of agents to reach perfect performance on the training set is unimportant for the present study; we suspect that the agent's comparatively small convolutional network fails to perceive important distinctions between the more intricate ShapeNet models.

\begin{table*}[!h]
\small \centering
\begin{tabular}{l|l|l|l|l}
\textbf{Template lang.} & \textbf{D.O.}                         & \textbf{I.O.}                   & \textbf{D.O. \& I.O.}                   &  \\ 
\textbf{(training)} & \textbf{synonym}                         & \textbf{synonym}                   & \textbf{synonym}                   & \textbf{Natural instruction} \\\hline
Put a mug    & Put a \underline{cup}    & Put a mug    & Put a \underline{cup}         & \textit{place mug in the basket}          \\
on the tray                                          & on the tray                                     &    on the \underline{box}                                       &        on the \underline{box}                                        & \textit{Keep the cup in atub}            \\
                                           &                                              &                                          &                                               & \textit{place the mug in a container}    \\
                                          &                                              &                                          &                                               & \textit{put the coffee mug in the box}    \\ \hline
Put a train   & Put a \underline{locomotive} & Put a train on &Put a \underline{locomotive} & \textit{Put the tractor on the bed}       \\
on the bed                                      &    on the bed                                 &      the \underline{bunk}                                    &      on the \underline{bunk}                                          &\textit{Move the train toy onto the bed}  \\
                                          &                                              &                                          &                                               & \textit{Place a toyvehicle on the bed}    \\
                                          &                                              &                                          &                                               & \textit{place the rail on tthe bed}      
\end{tabular}
\caption{\label{put_data} Example training (left) and test instructions in the putting task. Underlined words are examples of synonyms, italics indicate entire phrases provided by human annotators.}
\end{table*}

We consider two evaluation settings. In the \textbf{synonym} evaluation, we ran the environment for 1,000 episodes with the noun in the environment template instruction replaced by a synonym (\instr{Find a} $X$ becomes \instr{Find a} $X^*$ where $X \approx X^*$). The synonyms were provided by native English speaking subjects. In the \textbf{natural reference} evaluation, we gave 40 annotators access to a room containing a single ShapeNet model via a crowd-sourcing platform. We asked them to write down what they found in the room and then hit a button that restarted the environment with a new object in the room.\footnote{See Supplement for the all synonyms and human instructions.} We used this interactive method of soliciting instructions, rather than paraphrases of template instructions, because the setting more faithfully reflects the perspective of a user instructing a robot or situated learning system. 

As illustrated in Table~\ref{tab:liftexp}, unlike the synonyms, the natural referring expressions involve variation in articles as well as nouns (\emph{a pencil} might become \emph{the pen}), may include spelling mistakes or typos, can refer entirely incorrectly to the intended object (if the subject fails to recognize the ShapeNet model), but may also match the training instruction exactly. Moreover, unlike the synonym test, there are $30 - 40$ natural referring expressions for each of the 80 environment nouns, from which we sample randomly when evaluating the agent (again on 1,000 evaluation espisodes). Overall, the human referring expressions are highly varied; while there are 82 unique word types across all possible template instructions for this task, the human referring expressions involved 557 unqiue word types.The full set of human instructions aligned with ShapeNet model IDs can be downloaded from \url{https://tinyurl.com/s6u5bbj}

\subsection{Putting objects on other objects}

The strength of models like BERT is their ability to combine lexical representations into phrasal or sentence representations. To study this capacity in the context of instruction-following, we devised a \textbf{putting task} involving the verb `to put', which, in the imperative (\instr{put the cup on the tray}) takes two arguments, the \emph{direct object (D.O.)} \instr{cup} and the \emph{indirect object (I.O.)} \instr{tray}. In terms of the behaviours required, the putting task focuses on control and object relations rather than object identification or reference. The environment was configured to begin each episode with three randomly-chosen moveable objects and two larger immovable objects (a bed and a tray), each randomly positioned in the room. In each episode of this task, the agent receives an instruction \instr{Put a D.O. on the I.O.}, where \instr{D.O.} is any of the three moveable objects (chosen from a global set of ten) and \instr{I.O.} is either \instr{bed} or \instr{tray}. The environment checks whether an instance of \instr{D.O.} is at rest (and not held by the agent) on top of the \instr{I.O.}, returning a positive reward$ +1$ if so and ending the episode. If the object \instr{D.O.} is placed on something other than a \instr{I.O.}, or if another movable object is placed on the bed or the tray then again the episode ends immediately with reward $0$.\footnote{We found that ending the episode in such cases made learning much faster.}    

As before, we first trained all agents on the putting task with synthetic environment language instructions. Training a policy on this task with reinforcement learning required a bespoke task curriculum (see Supplement for more details); a well-trained policy completes each episode in an average of $\approx 50$ actions/timesteps. To gather the evaluation stimuli, we again crowd-sourced humans to provide both natural synonyms for each of the 12 objects in the global set for this task and, in this case, entirely free-form natural human instructions. To obtain natural instructions, we instantiated an environment with only one of the global set of moveable objects, coloured red, and one of either the bed or the tray, coloured white, and asked subjects to \emph{ask somebody to place the red object on top of the white object without mentioning their color}. This resulted in a set of natural instructions containing a total of 180 unique word types, compared to only 19 in the template instructions used for agent training. We then evaluated agents on four specific evaluations, illustrated in Table~\ref{put_data}: \textbf{D.O. synonym}, \textbf{I.O. synonym} and \textbf{D.O. \& I.O. synonym}, in which particular parts of the original template command were replaced with synonyms, and \textbf{Natural instruction}, the fully free-form human instruction, which can include orthographic errors and misidentified objects.

\subsection{Discussion of results}

\begin{table*}[!ht]
\centering
\small
\begin{tabular}{lccc}
\hline
 &\textbf{Template language} &  & \textbf{Natural} \\
\textbf{Model} &\textbf{(training)} & \textbf{Synonym} & \textbf{reference} \\
\hline
Random `lifting' act & 0.5 & 0.50 & 0.50  \\
Random-embedding + MP & $0.86 $ & $ 0.49  $ & $ 0.48  $ \\
Word-level Transf. & $ 0.91  $  & $ 0.58  $ & $ 0.61  $ \\
WordPiece Transf.  & $ 0.89  $ & $ 0.57  $ & $ 0.66  $ \\
Word-level Transf. + TN & $ 0.86  $  & $ 0.62  $ & $ 0.70  $ \\
WordPiece Transf. + TN  & $ 0.86  $ & $ 0.64  $ & $ 0.69  $ \\
\underline{Word embeddings} + Transf. & $ 0.89 $  & $ 0.62 $ & $ 0.64  $ \\
\underline{BERT} + MP & $ 0.93  $ & $ 0.77  $ & $ 0.76  $ \\
\underline{BERT} + SA & $ 0.91  $ & $ 0.70  $ & $ 0.70 $ \\
\underline{BERT} + CMSA & $ 0.88  $ & $ 0.68  $  & $ 0.67  $ \\
\underline{BERT} + CMSA + TN & $ 0.89  $  & $ 0.74  $  & $ 0.73  $ \\
Human performance estimate & --  & -- & $ 0.75 $\\
\hline
\end{tabular}
\caption{Accuracy on training instructions, the synonym evaluation and the natural referring expression evaluation in the reference task. MP: mean pool, SA: self-attention layer, CMSA: cross-modal self-attention layer, TN: typo-noise. Scores show mean across 1,000 episodes (with instructions randomly chosen by the environment generator).}\label{tab:liftresults}
\end{table*} 

\begin{table*}[h]
\centering
\small
\begin{tabular}{lccccc}
\hline
 & \textbf{Template lang} & \textbf{D.O.} & \textbf{I.O.} & \textbf{D.O. \& I.O.} & \textbf{Natural} \\
\textbf{Model} & \textbf{(training)} & \textbf{synonym} & \textbf{synonym} & \textbf{synonym} & \textbf{instruction} \\
\hline
Random `putting' act & 0.17& 0.17& 0.17& 0.17& 0.17\\
Random-embedding + MP & $ 0.99 $ & $ 0.59 $  & $ 0.22 $ & $ 0.15 $  & $ 0.39 $ \\
Word-level Transf. & $ 0.99  $  & $ 0.48  $  & $ 0.09  $ & $ 0.03 $  &$ 0.36 $\\
WordPiece Transf.  &  $ 0.98  $ & $ 0.30  $ & $ 0.24 $ &$ 0.10  $  &$ 0.42 $ \\
Word-level Transf. + TN & $ 0.97  $  & $ 0.61  $  & $ 0.27  $ & $ 0.17 $  & $ 0.45 $\\
WordPiece Transf. + TN  &  $ 0.97  $ & $ 0.36  $ & $ 0.36 $ &$ 0.13  $  &$ 0.56 $ \\
\underline{Word embeddings} + Transf.  & $ 0.99 $ & $ 0.57 $ & $ 0.47 $  & $0.36 $ & $ 0.55 $ \\
\underline{BERT} + MP  & $ 0.99 $ & $0.94 $ & $ 0.74 $ & $0.57 $ & $ 0.49 $ \\
\underline{BERT} + SA  & $ 0.99 $  & $ 0.96 $ & $ 0.69 $  & $0.56 $ & $0.54  $ \\
    \underline{BERT} + CMSA  & $ 0.99 $ & $ 0.93 $ & $ 0.67 $ & $0.47 $  & $ 0.43  $ \\
\underline{BERT} + CMSA + TN  & $ 0.98 $  & $ 0.88 $ & $ 0.79 $   & $ 0.70 $ & $ 0.70 $\\
Multitask \underline{BERT} + CMSA + TN  & $ 0.98 $  & $ 0.96 $ & $ 0.75 $   & $ 0.68 $ & $ 0.66 $\\
Human performance estimate & --  & -- & -- & -- & $ 0.69 $\\
\end{tabular}
\caption{Accuracy of different models on the putting task when different parts of instructions of the form \instr{"Put a [D.O.] on the [I.O.]"} are replaced with synonyms. Underlined words in model names indicate pre-trained weights from text-based training. MP: mean pool, SA: self-attention layer, CMSA: cross-modal self-attention layer, TN: typo-noise. Multitask: single agent trained on both reference and putting tasks. The evaluation task covers episodes in which \instr{[I.O.]} is either `bed' or `tray'. Scores show mean across 1,000 random episodes.}\label{tab:putresults}
\end{table*} 

The accuracies for the reference task are presented in Table~\ref{tab:liftresults} and for the putting task in Table~\ref{tab:putresults}. A video of the BERT+CMSA+TN agent both succeeding and failing when following human instructions on the putting task can be seen at \url{https://tinyurl.com/uy4fus2}. The results reveal the following main effects of language encoding on model performance:

\textbf{Substantial transfer from text requires contextual encoders} Agents with weights that are pretrained on text data exhibit substantially higher accuracy on both the reference and the putting tasks. This effect is greatest in the more focused synonym evaluations, but also holds for the the free-form human instructions. A small transfer effect can be seen by comparing the~\emph{word embeddings + Transformer condition} (62\% accuracy on the synonym evaluation, reference task and 57\% accuracy on the D.O. synonym evaluation, putting task) with the~\emph{WordPiece Transformer} (57\% and 20\%). However, overall the transfer effect is much stronger in the case of the full context-dependent BERT representations. On the same two evaluations, \emph{BERT + mean pool} achieves 77\% and 94\% accuracy respectively. The gains from transferring via BERT representations vs. just (sub)word embeddings are greatest for the (longer) putting instructions than for the reference (finding) instructions, and greatest of all in the \emph{D.O \& I.O. synonym evaluation}. These are cases where one would expect the marginal value of powerful sentential (rather than just lexical) representations should be greatest. 

\begin{figure}
    \centering
    \includegraphics[height=3.5cm,keepaspectratio]{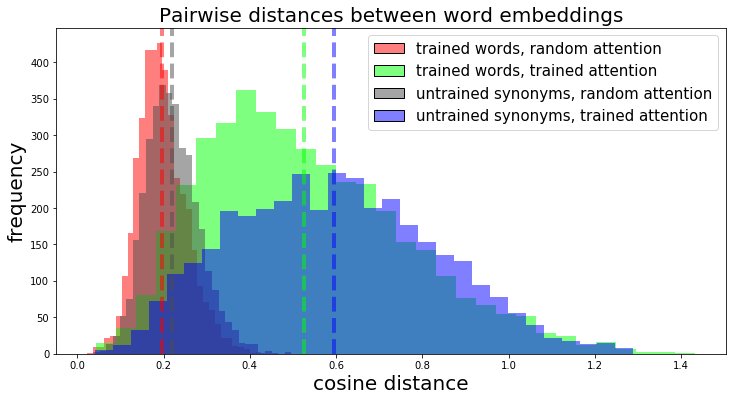}
    \caption{\label{fig:cosine} \footnotesize{Self-attention layers learn to pull-apart both training nouns \emph{and their synonyms} as the agent learns the putting task with synthetic environment language.}}
\end{figure}

\textbf{Tuning via self-attention layers (with typo noise) helps}
Interestingly, we find that tuned self-attention layers do not improve generalization performance over using BERT and mean pooling. This may be simply because the additional layers cause a degree of overfitting to the template environment language during training. However, typo noise mitigates this issue, so that the strongest evaluation performance on the putting tasks overall is observed with a combination of a tuned cross-modal self-attention layer and typo-noise training.\footnote{See Supplement for a comparison of BERT-based architectures with and without typo-noise.} Indeed, the value of typo noise as a regularizer can be seen by the fact that it improves the robustness of agents with tuned self-attention layers even in the \emph{synonym} evaluations (for both reference task and putting), which do not involve any typos. Thus, the \emph{BERT + CMSA + TN} model performs better than all others on two of the three synonym evaluations in the putting task. 

One way in which the appropriately-tuned self-attention layer might make the agent more robust to synonyms in this case is by spreading out task-relevant object-nouns in the agent's language representation space (leaving those words closer to synonyms than to potential confounding words). The degree to which this happens when the object-nouns and their synonymns in our environments are passed first through BERT and then through a (BERT + SA) agent-tuned self-attention layer (compared to passing through the same layer but with random weights) is shown in Fig~\ref{fig:cosine}. 

\textbf{WordPiece tokenization adds robustness} In the two evaluations involving natural language instructions from humans, a comparison of \emph{Word-level Transformer} and \emph{WordPiece Transformer} shows that some robustness is obtained simply from WordPiece encoding, which in turn must play some part in all BERT-based conditions (e.g. improving from $61$\% to $66$\% in the natural referring expression evaluation and from $36$\% to $42$\% in the natural instruction putting evaluation). As mentioned above, BERT-based agents with WordPiece encoding are particularly robust to human instructions when trained with typo noise, and this is most effective when combined with tuned self-attention layers. This makes intuitive sense, as a learnable self-attention layer should provide the agent with more flexibility to learn to correct for typos during training. Indeed, on the natural instruction evaluation of the putting task, where typos or spelling errors are most common, the cross-modal attention agent trained with typo noise achieves 70\% accuracy. Note that 100\% accuracy on this evaluation may be impossible, even for humans, because visual ambiguity or human error mean that instructions can sometimes refer in entirely mistaken ways to the objects in the room.  

\textbf{Scaling to multiple tasks} When evaluated on the putting task, the performance of an agent trained on both the putting and reference tasks is not substantially lower than agents that are specialized to each of the tasks individually.

\textbf{Agent performance on this data is close to ceiling} The best-performing agent achieves a similar level of performance to humans that are otherwise unfamiliar with the game or the instruction-setters. This underlines the extent to which the instructions procured from our human raters may be ambiguous or ill-formed. However, it also suggests that the best agents must be performing close to perfectly on all test episodes in which they have a reasonable chance to do so.

\section{Related work}

Most closely related to our work is an experiment by \citet{chan2019actrce}, who showed how an agent trained with \emph{InferLite} sentence representations~\citep{kiros2018inferlite} can be robust to synonym replacements in template instructions. The task itself involves object identification in the VizDoom environment~\citep{kempka2016vizdoom}, which requires only $3$ motor actions. Our work develops this insight substantially, applying a similar approach to a visually-realistic environment requiring 4, integrating context-dependent pre-trained models with subword tokenization (BERT), analysing architectures and training strategies for integrating such models and extending from synonym replacements to free-form instructions typed by humans. 

Much recent work applies deep learning and policy optimization in end-to-end approaches to learning instruction following~\citep{chaplot2018gated,oh2017zero,bahdanau2018learning,chevalier2018babyai,yu2018interactive,yu2018guided,jiang2019language}. As noted in the introduction, many of these studies, particularly those requiring policies to make fine-grained movements of the agent's body or objects, do not involve human language. 

Vision and language navigation (VLN) models learn to follow natural language directions that are longer that those considered in this work \cite{misra2017mapping,misra2018mapping,anderson2018vision,fried2018speaker,wang2019reinforced,zhu2019vision}. The best approaches to VLN do not apply deep RL methods, since VLN typically requires a high degree of exploration, rather than precise object control. VLN agents also often make use of privileged (i.e. non-first-person) observations~\cite{misra2018mapping}, knowledge of shortest paths ~\cite{fried2018speaker} and/or expert trajectories~\cite{wang2019reinforced}. In contrast, because we seek to evaluate a method that may ultimately be applied to robotics, we consider a setting in which the agent does not have access to privileged observations and/or gold-standard trajectories and an environment where different everyday objects must be controlled with a finer-grained set of actions ($4$ DoF). Another important difference is that, SHIFTT is a method for zero-shot transfer from template-based to natural instructions, whereas VLN models are both trained and tested on natural language instructions.

A limitation of all studies above, including this one, is the reliance on simulated environments. Both object identification and manipulation are likely far harder in reality, and it remains to be seen whether our proposed method would work seamlessly for robot language understanding~\citep{tellex2011approaching,tellex2012probabilistic,walter2014framework}. However, we note that it can in theory be applied in any multi-task or language-dependent policy-learning setting. See also~\citep{anderson2018vision,wang2019reinforced} for recent improvements to visual realism in simulated environments.

 Finally, there is a long history of building in knowledge about the structure of language and/or its environment into instruction-following systems rather than learning it end-to-end. In \cite{winograd1972understanding}'s SHRDLU, syntactic modules parsed the language input  into a logical form, and hand-written rules were applied to connect such forms to the environment. More recent pipeline-based approaches use learning algorithms to map language to a program that can then interface with a planner \citep{chen2011learning,matuszek2013learning,wang2016learning}, and/or a controller, both of which may have priviledged information about how the world connects to the program. It is likely that pretrained language encoders could add robustness to parts of these approaches, much as they do here. Our focus on end-to-end learning, however, is motivated by the intuition that it may eventually scale or adapt more flexibly to arbitrary environments or problems than pipeline approaches.

\section{Conclusions}

In this work, we have developed an agent that can follow natural human instructions requiring the identification, control and positioning of visually-realistic assets. Our method relies on zero-shot transfer from template language instructions to those given by human annotators when asked to refer and instruct in natural ways. The results show that, with powerful pretrained language encoders, this transfer effect is sufficiently strong to permit decoding extended language-dependent motor behaviours, despite the shift in distribution of the agent's input. More generally, we hope that this contribution serves to bring research on text-based and situated language learning closer together. To facilitate this, we make available our dataset of natural instructions and referring expressions aligned to ShapeNet models. 

While this study can be considered an interesting proof of concept for the SHIFTT approach to training linguistic deep RL agents via transfer-learning, there are many ways in which it can be improved and extended. As agents become better able to learn a wide range of conditional policies covering a larger set of motor behaviours, it will be instructive to scale the linguistic scope of the agent via the proposed technique, for instance to language involving verbs, or to questions and dialogue. Moreover, an important long-term objective is to apply SHIFTT to the training of robotic agents. There are also many alternative possibilities effecting the knowledge transfer from language model to agent that are not considered here. For instance, our approach involves freezing rather than fine-tuning the BERT encoder weights to our desired behaviour policy, to avoid overfitting, but techniques such as knowledge distillation~\citep{hinton2015distilling} could point to more elegant ways to learn jointly from text and environmental experiences. In addition, we have focused on BERT, but improvements may be possible by applying alternative general-purpose language encoders, such as GPT-2~\citep{radford2019language}, Roberta~\citep{liu2019roberta} and Transformer XL~\citep{dai2019transformer}. 

In sum, we have presented a conceptually simple recipe for transferring knowledge from a text corpus to a reinforcement-learning agent, and shown that the method permits zero-shot transfer from simulated to (constrained) natural language with surprising efficacy. We hope this opens new channels for research combining unsupervised (or semi-supervised) representation-learning with reinforcement learning, particularly at the intersection of language, vision and behaviour. 

\bibliography{paper}
\bibliographystyle{plainnat}

\section{Supplementary Material}

\subsection{Agent architecture and training details}

\begin{figure}[ht]
    \centering
    \includegraphics[height=8cm]{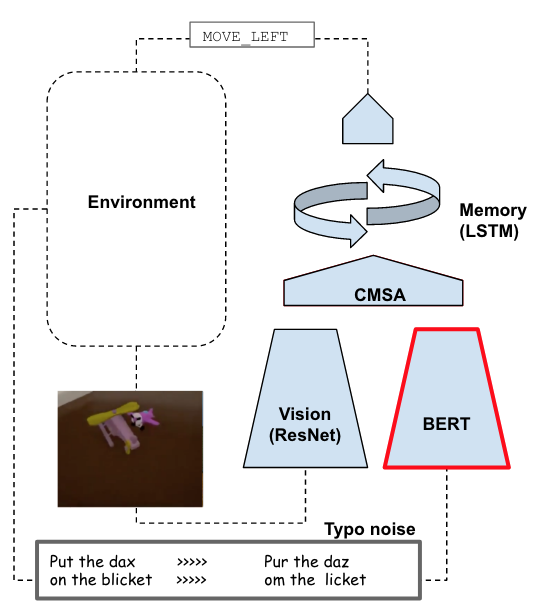}
    \caption{Schematic of the agent architecture with CMSA layer, and typo-noise applied to the environment's template instructions.}
\end{figure}

\label{app:agent}
\begin{table*}[t]
\begin{tabular}{@{}lll@{}}
\toprule
\textbf{Body movement actions} & \textbf{Movement and grip actions} & \textbf{Object manipulation}  \\ \midrule
NOOP                           & GRAB                               & GRAB + SPIN\_OBJECT\_RIGHT    \\
MOVE\_FORWARD                  & GRAB + MOVE\_FORWARD               & GRAB + SPIN\_OBJECT\_LEFT     \\
MOVE\_BACKWARD                 & GRAB + MOVE\_BACKWARD              & GRAB + SPIN\_OBJECT\_UP       \\
MOVE\_RIGHT                    & GRAB + MOVE\_RIGHT                 & GRAB + SPIN\_OBJECT\_DOWN     \\
MOVE\_LEFT                     & GRAB + MOVE\_LEFT                  & GRAB + SPIN\_OBJECT\_FORWARD  \\
LOOK\_RIGHT                    & GRAB + LOOK\_RIGHT                 & GRAB + SPIN\_OBJECT\_BACKWARD \\
LOOK\_LEFT                     & GRAB + LOOK\_LEFT                  & GRAB + PUSH\_OBJECT\_AWAY     \\
LOOK\_UP                       & GRAB + LOOK\_UP                    & GRAB + PULL\_OBJECT\_CLOSE    \\
LOOK\_DOWN                     & GRAB + LOOK\_DOWN                  &                               \\ \bottomrule
\end{tabular}
\caption{Agent action spec}
\label{app:action_set}
\end{table*}
The action set of the agent is presented in Table~\ref{app:action_set}.

To process visual input, the agent uses a residual convolutional network with $64, 64, 32$ channels in the first, second and third layers respectively and $2$ residual blocks in each layer.  

As the agent learns, actors carry replicas of the latest learner network, interact with the environment independently and send trajectories (observations and agent actions) back to a central learner. The learning algorithm~\citep{espeholt2018impala} modifies the learner weights to optimize an actor-critic objective, with updates importance-weighted to correct for differences between the actor policy and the current state of the learner policy. We used a learning rate of $0.0005$, a learner batch size of $32$, an agent unroll length of $50$, a discounting factor of $0.99$, an epsilon (epsilon-greedy policy) of $1e^{-6}$ and an \emph{entropy cost} of $ 0.0003$ ~\citep{mnih2016asynchronous}. We use an Adam optimizer~\citep{kingma2014adam} with $\beta_{1} = 0.90$ and $\beta_{2} = 0.95$.

In order to train the agent on the \emph{putting task}, it was necessary to combine episodes of the task itself with episodes of simpler tasks, in a form of curriculum (although learning in parallel on all tasks). In particular, we trained it concurrently on a \emph{reference task} that involved the same moveable objects as the putting task (so that the agent could receive signal about their names without relying on a complete act of putting). We also found it beneficial to add a \emph{put-near} task to the curriculum, where the instructions were of the form~\instr{Put a cup near the tray} rather than~\instr{Put a cup on the tray}, and a reward was emitted if the agent moved the cup a short distance from the tray and placed it on the ground. During training, the agent received equal experience of the modified reference task, the put-near task and the (target) putting task, and continued learning until performance on the putting task converged. When training the agent on the \emph{reference task}, no curriculum was required. When training the multi-task agent on both tasks, the agent was trained with equal experience on four tasks: the reference task, the modified reference task, the put-near task and the putting task. Training was stopped when performance on the putting task and the reference task converged. Note that the the distributed (IMPALA) framework lends itself naturally to sharing learning experience across tasks in this way, since we can simply have some proportion of the total agent actor threads interacting with particular tasks. The experience across the tasks in the aggregated into batches on the learner according to these proportions.  

For the \emph{reference task}, training the agent requires approximately 200 million frames of experience (approximately 7 million episodes), which takes about 24 hours with 250 actors on GPU. In the \emph{putting task}, training in each condition was stopped after 30 million episodes, approximately three days of training.

\subsection{Full experimental results}
\begin{figure*}[h]
    \centering
    \includegraphics[height=8cm]{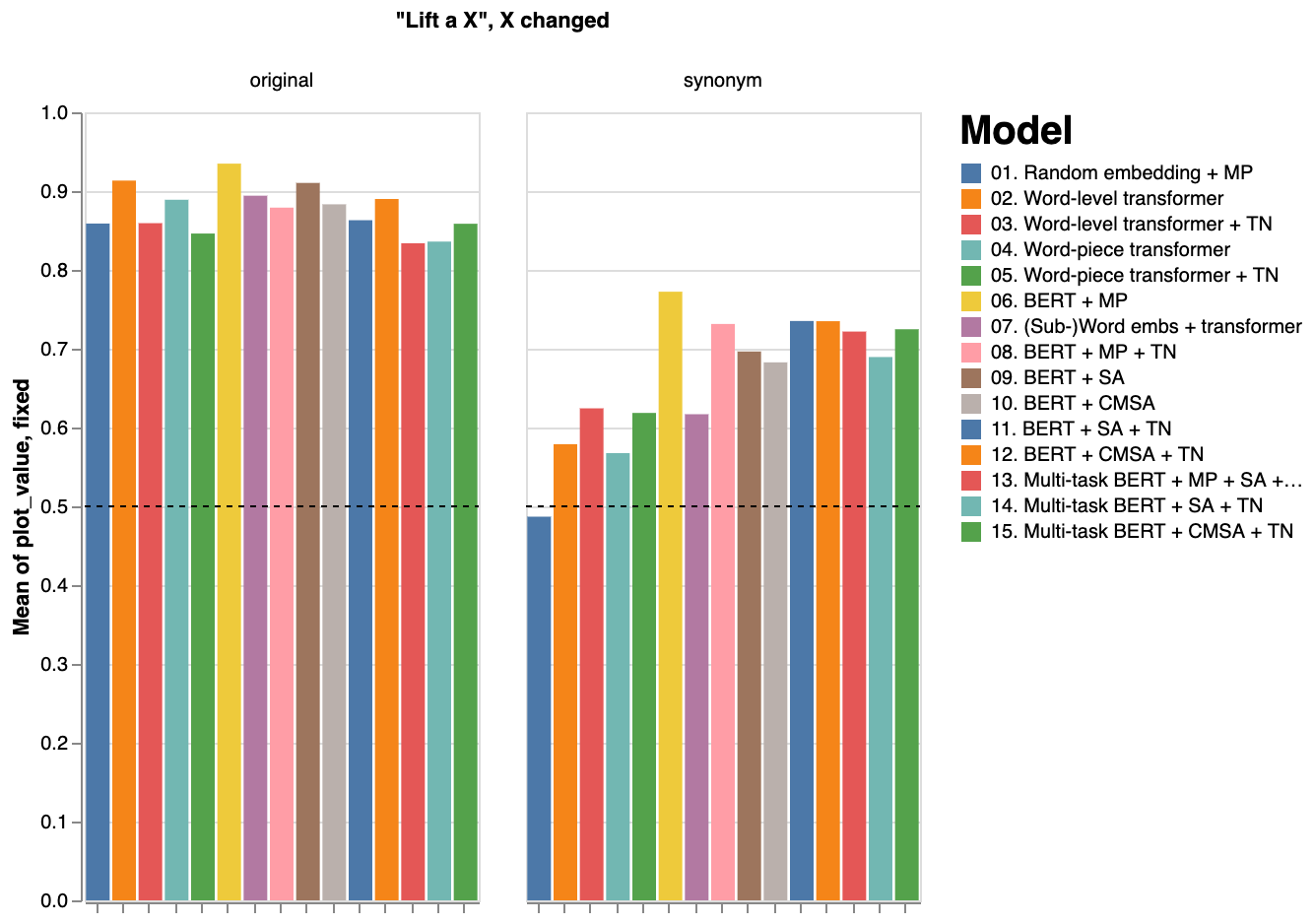}
    \caption{\textbf{reference task, synonym evaluation}. Left: accuracy of different agents over 1,000 episodes when no change is made to the environment template instruction. Right: accuracy of different agents when a synonym is introduced. Dotted line indicates score of an agent that carries out the correct behaviour but with random objects from the room.}
\end{figure*}

\begin{figure*}[h]
    \centering
    \includegraphics[height=8cm]{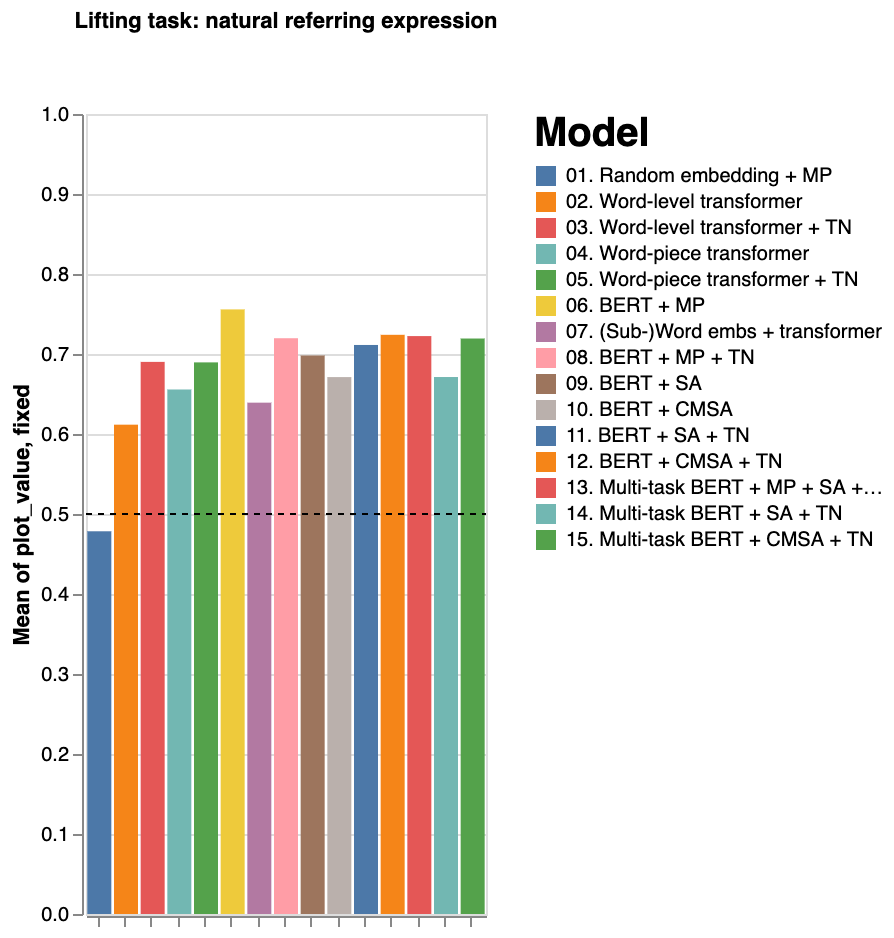}
    \caption{\textbf{reference task, natural referring expression evaluation}. Performance of agents on 1,000 evaluation episodes when part of the environment template reference task instruction is replaced by a natural referring expression from human annotators. Dotted line indicates score of an agent that carries out the correct behaviour but with random objects from the room.}
\end{figure*}

\begin{figure*}[h]
    \centering
    \includegraphics[height=7cm]{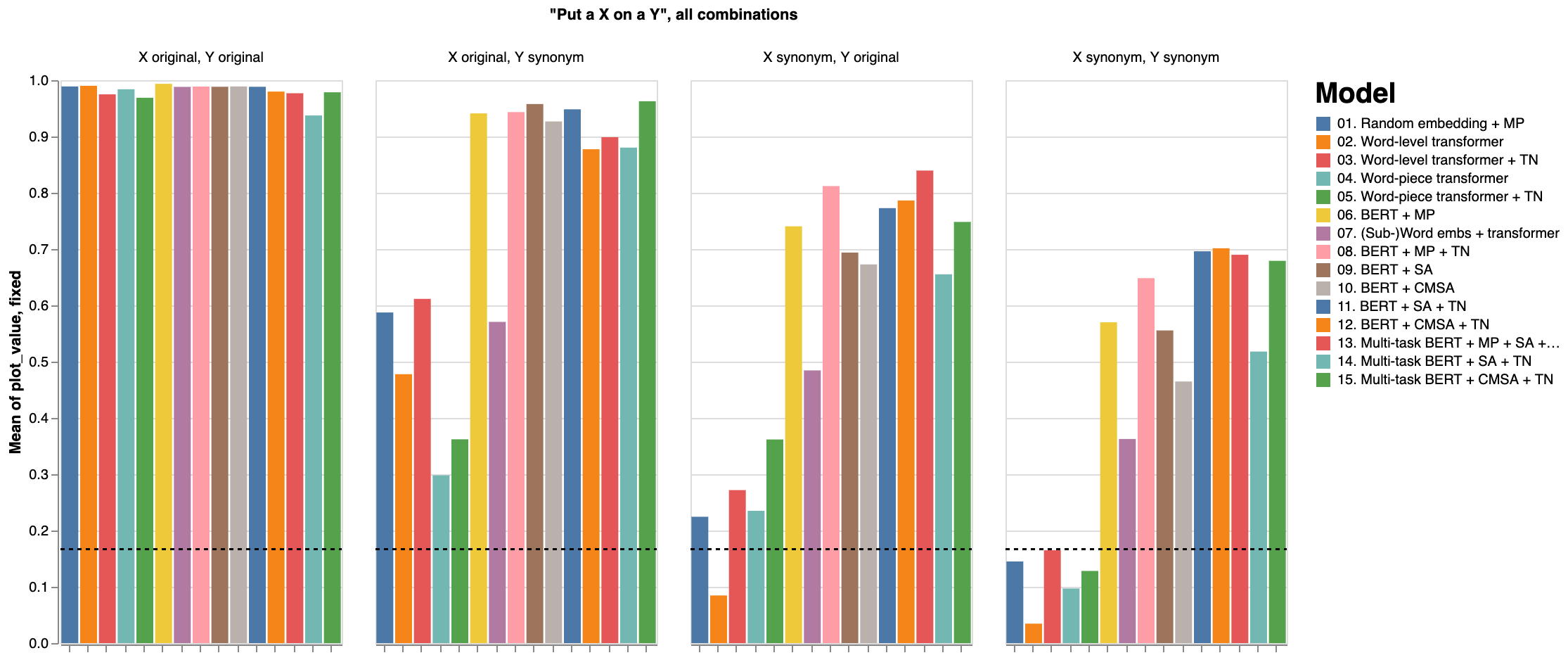}
    \caption{\textbf{Putting task, synonym evaluation}. Performance of agents on 1,000 evaluation episodes when (left) no change is made to the template environment instruction and (rightmost three) different parts of the template environment instruction are replaced with synonyms. Dotted line indicates score of an agent that carries out the correct behaviour but with random objects from the room.}
\end{figure*}

\begin{figure*}[h]
    \centering
    \includegraphics[height=8cm]{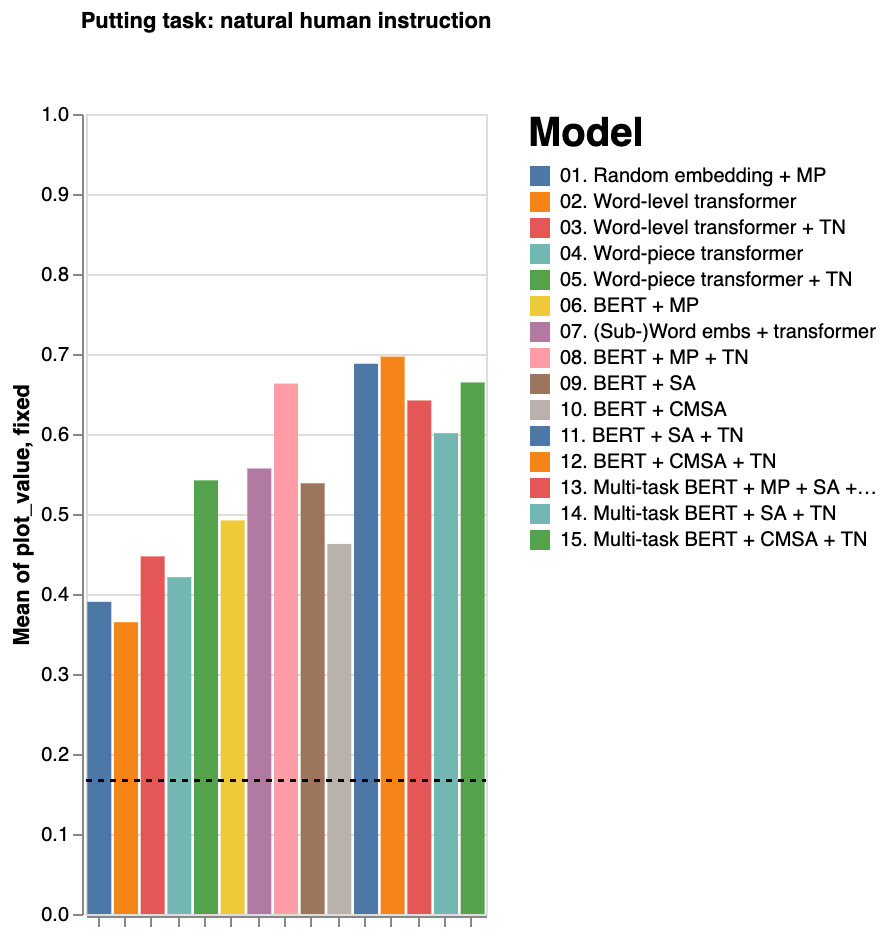}
    \caption{\textbf{Putting task, natural instruction evaluation}. Performance of agents on 1,000 evaluation episodes when the full environment template putting instruction is replaced by a natural instruction from human annotators. Dotted line indicates score of an agent that carries out the correct behaviour but with random objects from the room.}
\end{figure*}

\label{app:full_results}
For a superset of the experimental results presented in the main paper (with two additional conditions involving typo noise), see the Figures 1-4 on the following pages. 

\subsection{Instructions to human annotators}
\label{app:instructions}

\textit{Human annotators use a keyboard and mouse to control a player in the environment simulator, as is standard in first-person video games. The annotators were given the following instructions as part of the each annotation task.}

\subsubsection{Natural referring expressions}

This is a task called \textbf{Name The Object}. You will find yourself in a room containing a single object. Please move around the room to get a good view of the object. When you know what the object is:

\begin{enumerate}
    \item Hit Enter
    \item Type the name of the object
    \item Hit Enter again
\end{enumerate}

Examples of good responses:

\begin{enumerate}
    \item \textit{A kettle}
    \item \textit{Some trousers}
    \item \textit{A pair of scissors}
    \item \textit{A tennis ball}
\end{enumerate}

Please don't describe the object. Just write down what you see, with an article like `a' or `some' if appropriate. 

Example of bad responses

\begin{enumerate}
    \item \textit{ A brown ball}
    \item \textit{ A large piano with long black legs}
    \item \textit{ A small thing with lumps on the side}
\end{enumerate}

You should not need more than 4-5 different words, and most objects will require just 1 or 2 words to name.

Sometimes, you might be unsure what the object is. If that's the case, just make your best guess. 

\subsubsection{Full human instructions}

This is a task called \textbf{Ask to put}. You will find yourself in a room. Your job is to imagine giving an instruction to somebody else so that that person puts the red object on to the white object. Move around the room to get a good look at what the two objects are. When you are ready to give your instruction:

\begin{enumerate}
    \item Hit Enter 
    \item Type your instruction
    \item Hit Enter again
\end{enumerate}

Examples of good instructions:

\begin{enumerate}
    \item \textit{Place the cup onto the table}
    \item \textit{Put the ball on the plate}
    \item \textit{Move the pencil onto the box}
\end{enumerate}

\paragraph{Words to avoid} Your instruction must not contain words for colours or other properties of the object. Please do not use the words red, white, scarlet, dark, large etc. in your instruction. Instead, refer to objects by their name as you recognise them. If you don't recognise what an object is, just make your best guess. 

Examples of bad instructions:

\begin{enumerate}
    \item \textit{Put the red thing on the table}
\item \textit{Put the large object on the small object}
\item \textit{Put the small round thing on the chest}
\end{enumerate}

Keep your language varied. Try to use various ways to express your instruction in different episodes to keep things interesting.

\subsection{Further environment details}
\label{app:env}

For all experiments, the environment is a Unity room of dimension 4m x 4m. The walls, floor and ceiling are always the same color, but we add a window and door (positioned randomly per episode) to give some sense of absolute location to the agent. 

When importing ShapeNet models, we use the scaling provided in the metadata, unless it it is very small (less than 0.000001), in which case we interpret the coordinates in the OBJ file as meters. All objects have rigid bodies with collision meshes generated using Unity's built in MeshCollider, with convex set to true. The masses of all movable objects are set to 1 kg, so that our avatar has enough strength to pick all of them up (except for beds and trays, which are made kinetic).

When selecting ShapeNet models, for performance reasons we discarded all models with a vertex count higher than 8000, and an OBJ file size greater than 1 MB. The native ShapeNet category names are often not natural everyday names (for instance, they can be highly specific, like "dual shock analog controller"). To mitigate this, we used ShapeNet's WordNet tags, grouping models into categories according to WordNet synsets and assigning the name of the first synset lemma to the category. 

Because depth-perception is challenging without binocular vision, the agent is assisted in manipulation by a visual guide (bottom-right) that highlights objects within grasping range and drops a vertical line from held objects.

Full lists of objects and synonyms are on the following pages. 

\begin{table}[]
\scriptsize
\begin{tabular}{l|l|l|l|l}
\multicolumn{1}{l|}{\textbf{name}} & \multicolumn{1}{l|}{\textbf{synonym}} & \multicolumn{1}{l|}{\textbf{unique models}} & \multicolumn{1}{l|}{\textbf{example shapenet model id}} & \textbf{wordnet synset} \\ \hline
chest of drawers                             & cupboard                                        & 241                                                   & 793aa6d322f1e31d9c75eb4326997fae                                  & n3018908                          \\
table                                        & desk                                            & 199                                                   & db80fbb9728e5df343f47bfd2fc426f7                                  & n4386330                          \\
chair                                        & seat                                            & 167                                                   & 6bcabd23a81dd790e386ecf78eadd61c                                  & n3005231                          \\
sofa                                         & couch                                           & 158                                                   & 11d5e99e8faa10ff3564590844406360                                  & n4263630                          \\
television receiver                          & aerial                                          & 147                                                   & cebb35bd663c82d3554a13580615ae1                                   & n4413042                          \\
lamp                                         & light                                           & 138                                                   & 62ab9c2f7b826fbcb910025244eec99a                                  & n3641940                          \\
desk                                         & bench                                           & 98                                                    & a2c81b5c0ce9231245bc5a3234295587                                  & n3184367                          \\
box                                          & tin                                             & 81                                                    & 15b5c54a8ddc84353a5acb1c5cbd19ff                                  & n2886585                          \\
vase                                         & flute                                           & 78                                                    & 415a96bfbb45b3ee35836c728d324152                                  & n4529463                          \\
bed                                          & bunk                                            & 73                                                    & bbd23cc1e56a24204e649bb31ff339dd                                  & n2821967                          \\
floor lamp                                   & lamp                                            & 73                                                    & 54017ba00148a652846fa729d90e125                                   & n3371905                          \\
cabinet                                      & cupboard                                        & 68                                                    & f5024636f3514eb51d0662c550f8f994                                  & n2936496                          \\
book                                         & magazine                                        & 67                                                    & c8691c86e110318ef2bc9da1ba799c60                                  & n6422547                          \\
pencil                                       & pen                                             & 67                                                    & 839923a17af633dfb29a5ca18b250f3b                                  & n3914323                          \\
laptop                                       & computer                                        & 66                                                    & 20d42e934260b59c53c0c910fd6231ef                                  & n3648120                          \\
monitor                                      & screen                                          & 58                                                    & 4d11b3c781f36fb3675041302508f0e1                                  & n3787723                          \\
coffee table                                 & table                                           & 53                                                    & e21cadab660eb1c5c71d82cf1efbe60a                                  & n3067971                          \\
picture                                      & image                                           & 52                                                    & 2c31ea08641b076d6ab1a912a88dca35                                  & n3937282                          \\
bench                                        & chair                                           & 52                                                    & c6706e8a2b4efef5133db57650fae9d9                                  & n2832068                          \\
painting                                     & portrait                                        & 51                                                    & 6e51501eee677f7f73b3b0e3e8724599                                  & n3882197                          \\
rug                                          & carpet                                          & 44                                                    & b49fda0f6fc828e8c2b8c5618e94c762                                  & n4125115                          \\
shelf                                        & surface                                         & 43                                                    & 482f328d1a777c4bd810b14a81e12eca                                  & n4197095                          \\
switch                                       & flower                                          & 42                                                    & 7f20daa4952b5f5a61e2803817d42718                                  & n4379457                          \\
plant                                        & shrub                                           & 42                                                    & 94a36139281d1837465e08d496c0420f                                  & n17402                            \\
stool                                        & seat                                            & 41                                                    & 1b0626e5a8bf92b3945a77b945b7b70f                                  & n4334034                          \\
bottle                                       & jug                                             & 40                                                    & a429f8eb0c3e6a1e6ea2d79f658bbae7                                  & n2879899                          \\
loudspeaker                                  & speaker                                         & 37                                                    & 82b9111b3232904eec3b2e05ce8fd39b                                  & n3696785                          \\
electric refrigerator                        & fridge                                          & 36                                                    & 392f9b7a08c8ba8718c6f74ea0d202aa                                  & n3278824                          \\
toilet                                       & urinal                                          & 34                                                    & 5d567a0b5b57d8ab8b6558e44187a06e                                  & n4453655                          \\
dining table                                 & table                                           & 29                                                    & 82a1545cc0b3227ede650492e45fb14f                                  & n3205892                          \\
poster                                       & picture                                         & 28                                                    & de74ab90cc9f46af9703672f184b66db                                  & n6806283                          \\
wall clock                                   & clock                                           & 28                                                    & 1e2ea05e566e315c35836c728d324152                                  & n4555566                          \\
cellular telephone                           & mobile                                          & 26                                                    & 85a94f368a791343985b19765176f4ab                                  & n2995984                          \\
person                                       & human                                           & 25                                                    & 2c3dbe3bd247b1ddb19d42104c111188                                  & n5224944                          \\
cup                                          & mug                                             & 24                                                    & 542235fc88d22e1e3406473757712946                                  & n3152175                          \\
stapler                                      & clipper                                         & 24                                                    & 376eb047b40ef4f6a480e3d8fdbd4a92                                  & n4310635                          \\
mirror                                       & reflector                                       & 24                                                    & 36d2cb436ef75bc7fae7b9efb5c3bbd1                                  & n3778568                          \\
toilet tissue                                & toilet paper                                    & 24                                                    & 6658857ea89df65ea35a7666f0cfa5bb                                  & n15099708                         \\
desktop computer                             & pc                                              & 24                                                    & 102a6b7809f4e51813842bc8ef6fe18                                   & n3184677                          \\
table lamp                                   & desk light                                      & 23                                                    & 85b52753cc7e7207cf004563556ddb36                                  & n4387620                          \\
flag                                         & banner                                          & 22                                                    & ac50bd1ed53b7cb9cec8a10ad1c084eb                                  & n3359749                          \\
armchair                                     & settee                                          & 21                                                    & 806bce1a95268006ecd7cae46ee113ea                                  & n2741540                          \\
cupboard                                     & wardrobe                                        & 21                                                    & 3bb80aa0267a12bed00bff798ed59ff5                                  & n3152990                          \\
pen                                          & pencil                                          & 20                                                    & 628e5c83b1762320873ca101f05858b9                                  & n3913116                          \\
bag                                          & sack                                            & 20                                                    & b2c9ac70c58c90fa6dd2d391b72f2211                                  & n2776843                          \\
soda can                                     & drink                                           & 20                                                    & 16526d147e837c386829bf9ee210f5e7                                  & n4262696                          \\
sword                                        & knife                                           & 20                                                    & 555c17f73cd6d530603c267665ac68a6                                  & n4380981                          \\
bookshelf                                    & shelf                                           & 20                                                    & 586356e8809b6a678d44b95ca8abc7b2                                  & n2874800                          \\
fireplace                                    & burner                                          & 19                                                    & df1bd1065e7f7cde5e29ce2c9d37b952                                  & n3351301                          \\
curtain                                      & drape                                           & 19                                                    & 6f3da555075ec7f9e17e45953c2e0371                                  & n3155743                          \\
battery                                      & cell                                            & 18                                                    & 62733b55e76a3b718c9d9ab13336021b                                  & n2813606                          \\
bookcase                                     & bookshelf                                       & 18                                                    & bc80335bbfda741df1783a44a88d6274                                  & n2874241                          \\
pizza                                        & pie                                             & 18                                                    & caca4c8d409cddc66b04c0f74e5b376e                                  & n7889783                          \\
hammer                                       & mallet                                          & 17                                                    & a49a6c15dcef467bc84c00e08b1e25d7                                  & n3486255                          \\
microwave                                    & oven                                            & 16                                                    & b3bf04a02a596b139220647403cfb896                                  & n3766619                          \\
cereal box                                   & oatmeal                                         & 16                                                    & dc394c7fdea7e07887e775fad2c0bf27                                  & n3001610                          \\
food                                         & meal                                            & 16                                                    & d92f30d9e38cf61ce69bbdea737daae6                                  & n21445                            \\
screen                                       & monitor                                         & 16                                                    & 4d7fb62f0ed368a18f62bdf4e9082924                                  & n4159912                          \\
glass                                        & beaker                                          & 16                                                    & 89cf9af7513ecd0947bdf66811027e14                                  & n3443167                          \\
wine bottle                                  & magnum                                          & 15                                                    & e101cc44ead036294bc79c881a0e818b                                  & n4599016                          \\
lamppost                                     & streetlight                                     & 15                                                    & 9cf4a30ab7e41c85c671a0255ba06fe5                                  & n3642472                          \\
oven                                         & cooker                                          & 15                                                    & 69c57dd0f6abdab7ac51268fdb437a9e                                  & n3868196                          \\
camera                                       & polaroid                                        & 14                                                    & 48e26e789524c158e1e4b7162e96446c                                  & n2946154                          \\
globe                                        & world                                           & 14                                                    & 3e97d91fda31a1c56ab57877a8c22e14                                  & n3445436                          \\
cd player                                    & stereo                                          & 14                                                    & fe90d87deaa5a8a5336d4ad4cfab5bfe                                  & n2991759                          \\
wardrobe                                     & cupboard                                        & 14                                                    & cb48ec828b688a78d747fd5e045d7490                                  & n4557470                          \\
bible                                        & text                                            & 14                                                    & 6ad85ddb5a110664e632c30f54a9aa37                                  & n6434286                          \\
pillow                                       & cushion                                         & 13                                                    & 8b0c10a775c4c4edc1ebca21882cca5d                                  & n3944520                          \\
mug                                          & cup                                             & 13                                                    & ea33ad442b032208d778b73d04298f62                                  & n3802912                          \\
chandelier                                   & light                                           & 13                                                    & 37d81dd3c640a6e2b3087a7528d1dd6a                                  & n3008889                          \\
chessboard                                   & checkers                                        & 13                                                    & ec6e09bca187c688a4166ee2938aa8ff                                  & n3017971                          \\
calculator                                   & computer                                        & 13                                                    & 387e59aec6f5fdc04b836f408176a54c                                  & n2942270                          \\
dishwasher                                   & washing machine                                 & 12                                                    & ff421b871c104dabf37a318b55c6a3c                                   & n3212662                          \\
mattress                                     & mat                                             & 12                                                    & f30cb7b1b9a4184eb32a756399014da6                                  & n3736655                          \\
basket                                       & bag                                             & 12                                                    & 91b15dd98a6320afc26651d9d35b77ca                                  & n2805104                          \\
pencil sharpener                             & sharpener                                       & 12                                                    & 287b8f5f679b8e8ecf01bc59d215f0                                    & n3914833                          \\
candle                                       & lantern                                         & 12                                                    & cf1b637d9c8c30da1c637e821f12a67                                   & n2951508                          \\
bowl                                         & dish                                            & 12                                                    & 594b22f21daf33ce6aea2f18ee404fd5                                  & n2884435                          \\
clock                                        & watch                                           & 12                                                    & 299832be465f4037485059ffe7a2f9c7                                  & n3050242                          \\
coat hanger                                  & hanger                                          & 12                                                    & 5aa1d74d04065fd98a7103092f8e8a33                                  & n3061905                         
\end{tabular}
\caption{The 80 ShapeNet category names and their sizes, corresponding WordNet synsets and an example model id for a category exemplar used in the \emph{reference task}. In the extra materials we provide a full list of ShapeNet model ids in each category.}
\end{table}

\begin{table*}[!h]
\begin{tabular}{r|l|l}
\multicolumn{1}{l|}{}             & \textbf{Environment word} & \textbf{Synonym} \\ \hline
Immovable objects                 & tray                      & box              \\
                                  & bed                       & mattress         \\ \hline
\multirow{13}{*}{Movable objects} & boat                      & ship             \\
                                  & hairdryer                 & dryer            \\
                                  & racket                    & bat              \\
                                  & bus                       & coach            \\
                                  & rocket                    & spaceship        \\
                                  & car                       & automobile       \\
                                  & plane                     & aeroplane        \\
                                  & mug                       & cup              \\
                                  & robot                     & android          \\
                                  & train                     & locomotive       \\
                                  & keyboard                  & piano            \\
                                  & helicopter                & airplane         \\
                                  & candle                    & lamp            
\end{tabular}
\caption{The two immovable and ten movable objects in the putting experiment. To begin each episode, both immovable objects and three randomly-selected movable objects are randomly positioned in the room.}
\end{table*}


\newpage

\end{document}